\documentclass[preprint,3p,times]{elsarticle}

\usepackage{booktabs}
\usepackage{amsmath,amssymb,amsfonts}
\usepackage{algorithmic}
\usepackage{graphicx}
\usepackage{textcomp}
\usepackage{xcolor}
\usepackage{multirow}
\usepackage{hyperref}
\usepackage{subcaption} 
\def\BibTeX{{\rm B\kern-.05em{\sc i\kern-.025em b}\kern-.08em
    T\kern-.1667em\lower.7ex\hbox{E}\kern-.125emX}}
\usepackage{enumitem}
\usepackage{amssymb}
\usepackage{amsmath}

\usepackage{xcolor}
\newcommand{\MV}[1]{\textcolor{black}{#1}}

\journal{Expert Systems With Applications}

\begin{document}

\begin{frontmatter}



\title{BOOST: Out-of-Distribution-Informed Adaptive Sampling for Bias Mitigation in Stylistic Convolutional Neural Networks}

\author[lbl1]{Mridula Vijendran}
\ead{mridula.vijendran@durham.ac.uk}
\author[lbl1]{Shuang Chen}
\ead{shuangchen@durham.ac.uk}
\author[lbl1]{Jingjing Deng}
\ead{jingjing.deng@durham.ac.uk}
\author[lbl1]{Hubert P. H. Shum\corref{cor1}}
\ead{hubert.shum@durham.ac.uk}


\cortext[cor1]{Corresponding author}


\affiliation[lbl1]{organization={Department of Computer Science, Durham University}, 
             city={Durham}, 
             country={UK}}




\begin{abstract}
The pervasive issue of bias in AI presents a significant challenge to painting classification, and is getting more serious as these systems become increasingly integrated into tasks like art curation and restoration. Biases, often arising from imbalanced datasets where certain artistic styles dominate, compromise the fairness and accuracy of model predictions, i.e., classifiers are less accurate on rarely seen paintings. While prior research has made strides in improving classification performance, it has largely overlooked the critical need to address these underlying biases, that is, when dealing with out-of-distribution (OOD) data. Our insight highlights the necessity of a more robust approach to bias mitigation in AI models for art classification on biased training data. We propose a novel OOD-informed model bias adaptive sampling method called BOOST (Bias-Oriented OOD Sampling and Tuning). It addresses these challenges by dynamically adjusting temperature scaling and sampling probabilities, thereby promoting a more equitable representation of all classes. We evaluate our proposed approach to the KaoKore and PACS datasets, focusing on the model's ability to reduce class-wise bias. We further propose a new metric, Same-Dataset OOD Detection Score (SODC), designed to assess class-wise separation and per-class bias reduction. Our method demonstrates the ability to balance high performance with fairness, making it a robust solution for unbiasing AI models in the art domain.
\end{abstract}






\end{frontmatter}

\section{Introduction}
\label{intro}

Artificial Intelligence (AI) is becoming increasingly significant in the analysis of paintings. AI tools extract and analyze artworks for deeper insight, supporting art appraisers \cite{gonthier2018weakly} or visitors \cite{hutson2024integrating} to understand and appreciate extensive art collections. For instance, mathematical models quantify aspects like shape, shading, and cast shadows in realist drawings, aiding in the accurate representation and analysis of visual elements \cite{stork2006mathematical}.  These AI-based analyses facilitate the preservation of cultural heritage, ensuring a nuanced understanding of diverse cultures across different time periods and geopolitical contexts \cite{ghaith2024ai}. Through these applications, AI is transforming how we engage with and appreciate the rich history and diversity of painting.

Models trained on real-world datasets have been applied to painting analysis. By utilizing knowledge gained from large, diverse datasets, both off-the-shelf models and more specialized architectures can more efficiently adapt to the unique features of art, requiring less task-specific data while still delivering accurate and meaningful insights. Conventional Neural Networks (CNNs) excel at tasks such as scene categorization or object detection in artworks by adapting to the domain using techniques augmenting and emphasizing style elements \cite{rezanejad2019gestalt, Jeon2020, kadish2021improving}. However, these models often rely on the assumption that the training data is similar enough to the target artistic styles, which can lead to performance degradation when faced with significant domain gaps. To address this, some researchers have fine-tuned models specifically for highly stylized datasets, improving performance in tasks such as human pose estimation in ancient vase paintings \cite{PRATHMESH2022, Nicholas2016}. Mixed architecture models combine general feature extractors with domain-specific modules to enhance object detection through advanced learning techniques \cite{gonthier2018weakly,wang2023cut, Marinescu2020, Madhu2022}. Despite these advancements, a key limitation of existing work is that these models, even when adapted to the specific data distribution of artistic tasks, often introduce or fail to mitigate biases inherent in the data. Accuracy-focused improvements and architectural adaptations overlook biases in the art domain, which impact the interpretation and cultural significance of artworks.

Distinguishing and avoiding bias is a major challenge in the AI system used in real-world painting-related tasks \cite{zhang2022reducing}. In the real world, models typically contend with a consistent visual style on real-world content, but in the painting domain, there exists real and imaginary content with different artistic styles, ranging from classical paintings to modern abstract forms. This diversity often presents itself as the class imbalance in painting datasets, where certain contents and styles are much more common than others. During training, this imbalance leads to model bias as more frequent classes are prioritized over rarer ones, making the model better at recognizing dominant styles while being less effective at identifying uncommon or niche forms of painting. This limitation not only affects the accuracy of AI tools in recognizing diverse painting forms but also risks perpetuating existing cultural biases in the appreciation and understanding of painting.

\begin{figure}[ht]
\centering

   \includegraphics[width=0.7\linewidth]{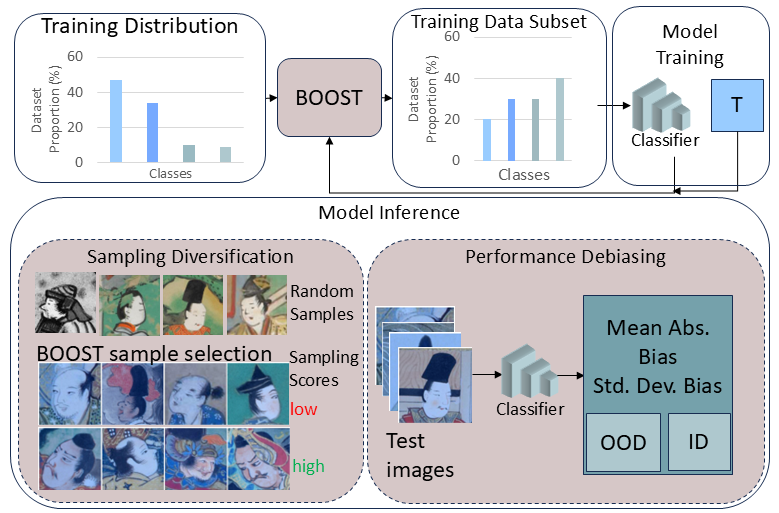}
    \caption[short for lof]{Overall architecture design for incorporating both BOOST data sampling and image classification pipelines.}
\label{fig:overall}

\end{figure}

The proposed ODIN (out-of-distribution detector for neural networks) \cite{liang2018enhancing}-informed sampling method (Figure \ref{fig:overall}) helps mitigate bias towards high-frequency classes in class-imbalanced datasets. By scaling the softmax scores with class-wise averages, the method calibrates rare and ambiguous samples to be closer to common samples, promoting a more inclusive representation of all classes. Additionally, the dynamic sampler temperature during training adapts the model to both inter- and intra-class distributions without strong inductive priors that average out important information in larger datasets.  We seek the intra-class distance minimization while maximizing the inter-class’ to reduce the number of ambiguous samples and improve the separation between classes respectively. Finally, the method prioritizes rarer classes and their harder samples by reversing the sampling probabilities and using multinomial batch stratification. This approach promotes the inclusion of underrepresented classes and challenging examples in the training process. The framework supports additional functionality in terms of samples diversification sorted according to the computed sampling scores corresponding to the learned model representation of the dataset.  We also introduce three metrics to measure biases across classes in common performance metrics, along with re-contextualizing OOD and ID samples between classes of the same dataset. Our OOD detection metric, SODC, enables a data distribution aware softmax profile to compute the deviation of sample prediction against the true class-wise distribution. We appreciate the reviewer’s feedback. Our Out of distribution (OOD) detection metric, SODC,  repurposes the OOD evaluation task to address multi-class prediction bias unlike other OOD detection metrics. Other OOD detection metrics use a single collapsable scalar score (like confidence \cite{techapanurak2021practical, hsu2020generalized, wang2010determining} or a distance \cite{techapanurak2021bridging, li2019repair, aleem2017inferring}) for binary separation, but we analyze the full softmax profile in relation to the training data's class distribution. Additionally our SODC promotes a semantic alignment with the class wise sampled images and the trained model along with a lower computational overload from using the classifier head only for computation. This structured alignment supports relative comparisons against the other classes and enhances data separability through deviations against the true class distribution.  The semantic alignment check allows for the detection of class-wise OOD samples whereas other detection metrics such as Mahalanobis provide this functionality at the feature space. Our model achieves a comparable performance of 84.44\% accuracy and an F1 score of 79.79\% in the KaoKore dataset. Our BOOST sampler has the lowest bias mean scores across all performance metrics.

Our code is available at: {\color{blue} \href{https://github.com/41enthusiast/BOOST}{https://github.com/41enthusiast/BOOST}}. The main contributions are as follows:  
\begin{itemize}[noitemsep,topsep=0pt]
    \item We present a solution to de-bias deep learning models against dominant data for fairer painting classification with a sampling method. This is further supported by a set of novel metrics to measure bias in both class-wise performance and out-of-distribution sample performance.
    \item We present a novel framework for de-biased image classification, involving a novel BOOST sampler and a classification backbone. The proposed BOOST sampler redefines out-of-distribution (OOD) and in-distribution (ID) samples to maximize inter-class differences, thereby facilitating de-biased downstream classification.
    \item We propose an adaptive temperature scheduler to learn sharper and more accurate decision boundaries by leveraging inter-class and intra-class separations from diversified learnable representations. This is achieved by providing subsets of the training data determined by both the training time scaled temperature and model confidence during sampling.
\end{itemize}

\section{Related Work}
\label{sec:related}

\subsection{OOD Detection}
In the literature on image classification with out-of-distribution (OOD), the following methods assess the generalization of classifiers to unseen classes not included in the training data distribution, focusing on maintaining classification performance on in-distribution tasks while adapting to OOD scenarios. In the problem of Open-set recognition where the model is trained on known classes and has to classify between known and unknown classes, prior research found correlations between the two classes' discriminability and models with high classification \cite{vaze2022openset}. While the work showcases further improvements in distributional shift and learned invariances with learning the two tasks successively, our work aims to use the learned attributes from the tasks to improve each other's performance. On the other hand, some studies find that models trained simultaneously on image classification and OOD detection suffer from trade-offs in performance \cite{techapanurak2021bridging, techapanurak2021practical} with larger degradations when using OOD samples as ambiguous, adversarial samples or highly corrupted images . This is largely attributed to changes in the network architecture or its training methods for solving OOD detection after the model is trained on image classification which is unapplicable to our architecture design where we use the OOD detection only in sampling data. Similar works use ODIN for input preprocessing and avoid tuning the model for the OOD task to reframe the problem setting to include both OOD and image classification, with the method highly effective in separating both distributional shifts and non-semantic shifts \cite{hsu2020generalized}.

\subsection{Bias Mitigation Techniques}
The examination of biases in visual art has evolved significantly, with recent studies focusing on more specific forms of bias. \cite{zhang2022reducing} discusses several types of biases such as sampling, imaging and evaluation biases that can occur in AI-based analysis of artworks and provides strategies to mitigate them. They find that Sampling bias can occur when the training data is skewed towards certain artists, styles, or time periods, the AI model may not generalize well to other types of artworks due to a lack of available data\cite{li2019repair, wang2010determining}. To reduce selection bias, the authors suggest using diverse and representative datasets that cover a wide range of artists, styles, and historical periods. They also recommend using data augmentation techniques to increase the variety of training examples. Depending on the curation, the label representations are biased towards different socio-cultural aspects. For example, \cite{surapaneni2020exploring} finds bias in gender labels due to human error and a lack of standardization in interpretation for the task of annotation which they mitigate by classifying the images to unassuming, generic attributes. Our work aims to address the resulting class-imbalance by unbiasing the model through increasing the interclass gap and oversampling ambiguous classes. Imaging bias can arise when the features or attributes used to describe artworks are not accurate or consistent. For instance, if the color or texture measurements are affected by the aging of paintings, lighting conditions or the quality of the digital images, it can lead to biased analysis results \cite{trumpy2015experimental,berns2016color}. The authors propose using standardized protocols for image acquisition and preprocessing to ensure consistency in measurements. They also suggest using multiple imaging modalities (e.g., visible light, infrared, X-Ray) to capture different aspects of the artworks. Other research find bias towards the artistic medium or geography due to accessibility and stakeholders \cite{wasielewski2022beyond}. Evaluation bias can occur when the performance of AI models is assessed using metrics or criteria that are not appropriate for the specific task or domain. For example, using accuracy as the sole evaluation metric may not capture the nuances of artistic style or the subjectivity of aesthetic judgments \cite{kao2017deep}. This could be mitigated through other metrics that measure the similarity of extracted features to annotations such as pose, segmentation maps capture variations under class labels. \cite{aleem2017inferring} explored artists' biases, measuring them according to fluency variables, in portrait paintings towards different poses depending on how they affect the overall symmetry, balance and complexity.

\subsection{Sampling Strategies}
The simplest form of adaptive data sampling, unrestricted random sampling \cite{ElKorchi2019UnrestrictedRS}, reduces the chance of a learning algorithm co-adapting with the training dataset by minimizing the reuse of the same samples in each iteration. They mitigate sampling bias by reducing co-adaptation to dominant patterns and minimizing repeated sample exposure during training. Random sampling alone does not prevent underrepresentation of minority classes during training. This is further excerbated in long-tailed distributions, where the absolute number of hard instances in the tail classes is still low compared to their proportion in head classes. The setting incentivizes class difficulty-weighted samplers \cite{sinha2022class}, which sample more frequently from difficult classes, computed based on both the cross-entropy loss and classifier rebalancing. But, weighing the sampling probabilities on instantaneous losses can lead to noisy or mislabeling of hard classes since their atypical samples can skew the model biases for rare classes. Additionally, the class difficulty metric is often computed using a fixed criteria and not tailored to the model's parameters. In noisy training data distributions with large proportions of anomalous data, research that use multiple classifier heads to represent different levels of difficulty of training subsets \cite{yoon2021self} degrade far as compared to contrastive learning methods. Their research is aimed towards removing noisy samples rather than improving class bias or addressing intra-dataset bias. This reduces the need of manual filtering of the training data, improving image classification on scrapped or streamed large unlabeled datasets. These models commonly need a pretraining stage before the post hoc OOD inference for robust and generalizable image classifiers with less importance to hyperparameter tuning, but come at the cost of long data sampling stages. They also require highly varied learned representations which could collapse into smaller modes after domain adaptation to a different problem domain. Our work gradually increases the variations of the learnable representations by providing subsets of the training data determined by the temperature scheduling at training and model confidence during sampling to prevent this modal collapse.

\begin{figure*}[ht]
\centering

   \includegraphics[width=.8\linewidth]{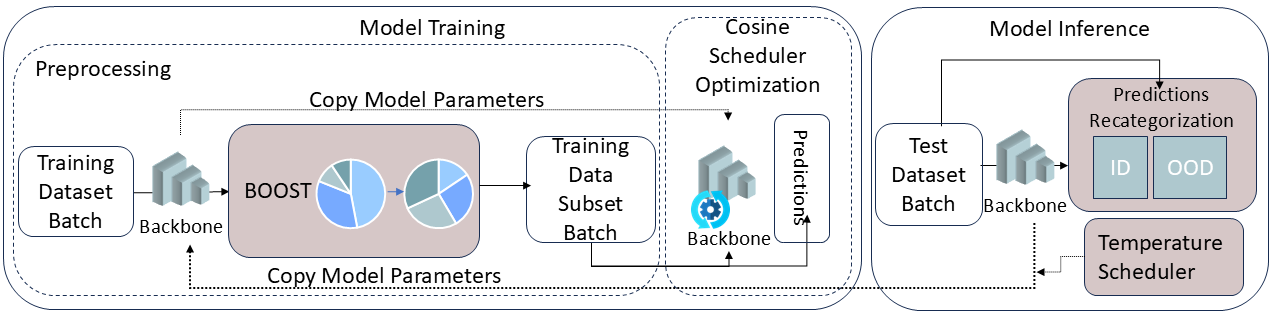}
\caption[short for lof]{Model training and inference pipeline with feedback to our sampler using the updated model and temperature scheduler. During the training phase, a batch of training data is first preprocessed and passed through a backbone model. The BOOST module then uses a copy of the model parameters to generate a refined training subset, which focuses on reducing class imbalance. In the inference phase, a batch of test data is passed through the backbone model, and predictions are recategorized into ID and OOD classes to compute the OOD score and compute class-separation and bias. A temperature scheduler is employed to fine-tune prediction confidence, improving separation between ID and OOD samples. This structured pipeline effectively enhances model performance by addressing both inter-class and intra-class imbalances and refining the decision boundaries.}
 \label{fig:model_train}
\end{figure*}
\section{Overview}

\label{sec:methodology}

We introduce a novel sampler-driven learning framework designed to avoid bias in multiclass image classification, particularly addressing challenges with class imbalance and context variations. We detail the task \& terms definition (sec.\ref{sec:task_terms_4.1}), our motivation (sec.\ref{sec:motivation4.2}) and the two-stage setup (sec.\ref{sec:training_testing_4.3}) in Sec:~\ref{sec:framework}. The overall framework (Figure \ref{fig:model_train}) is motivated by the observation that large class imbalance often causes models to overfit on dominant styles or classes, losing class-wise information. Our methodology addresses this by selecting representative samples from each class that exhibit high variation in the model space. This dataset selection prevents the classifier from forming fixed feature representations, ensuring a more generalizable learning process across all classes.

To implement this framework specifically for classification debiasing, we propose an adaptable BOOST sampler, integrated as a plug-and-play module that can seamlessly work with various model architectures. We present the design details in sec.~\ref{boost_sampler5.1}. BOOST sampler enhances bias robustness through strategic sampling, dynamic temperature scheduling, and alignment of logit and class distributions. We detail the engineering implementation of the proposed sampler. Finally, we propose a new metric in sec.~\ref{new_metrics5.2}, Same-Dataset OOD Detection Score (SODC), to evaluate the effectiveness in class-wise separation and bias reduction.

\section{A Sampler-Based Debiasing Framework} 
\label{sec:framework}
Our sampler-based debiasing framework is sample-driven, guiding the model's learning process to address both inter-class and intra-class imbalances through the adaptive selection of diverse training samples  as shown in Figure \ref{fig:model_train}. By re-sampling the training distribution, we can prevent the degradation of the majority class performance while improving the minority class performance in class-imbalanced datasets. Additionally, we employ a temperature scheduler to change the underlying sampler distribution to focus on learning hard samples to a uniform sampling distribution across epochs. During inference, we recategorize predictions based on misclassifications as ID and OOD, treating OOD samples as those hard examples that the model struggles to classify correctly, developing well-separated representations. The adaptive resampling and temperature scheduling together allow for attribute bias corrections within classes and majority bias between classes.

\subsection{Task and Terms Definition}
\label{sec:task_terms_4.1}
We introduce an overall pipeline to address class imbalances as learned by the model on datasets with unequal representations of classes. Skewed datasets have an uneven distribution of samples across classes, resulting in trained models having better performance on the majority classes at the cost of minority class samples. 

We categorize these class imbalances into inter-class imbalances and intra-class imbalances. Inter-class imbalances arise from uneven sample distributions between different classes, causing the model to be biased towards classes with more frequent samples. Intra-class imbalances, on the other hand, result from variations in task complexity, noisy or overlapping data, and rare instances within the same class, affecting the model’s ability to learn distinct patterns.

\subsection{The Motivation of Sampling-Based Debiasing}
\label{sec:motivation4.2}
The model-adaptive sampler introduces a controlled mechanism to manage the type of debiasing (intra- and inter-class levels) during training by adjusting the sampler temperature and model training hyperparameters. Painting datasets suffer from large intra-class imbalances and variability, particularly in diverse styles like Impressionism, which challenge classifiers \cite{zhao2021compare}. The group disparities in the dataset and the data sampling process directly impact the training dynamics, affecting how features are learned alongside the gradient components of different classes \cite{lampinen2024learned, ghosh2024class}. The model's feature representation space are biased towards the order of features learned by the model, implying that the order of samples introduced to the model affects the learned bias. Furthermore, majority class samples often exhibit larger gradient magnitudes, which in turn affect the weight update process and exacerbate inter-class bias in the model’s feature space.

\begin{figure*}[ht]
\centering

 \includegraphics[width=.95\linewidth]{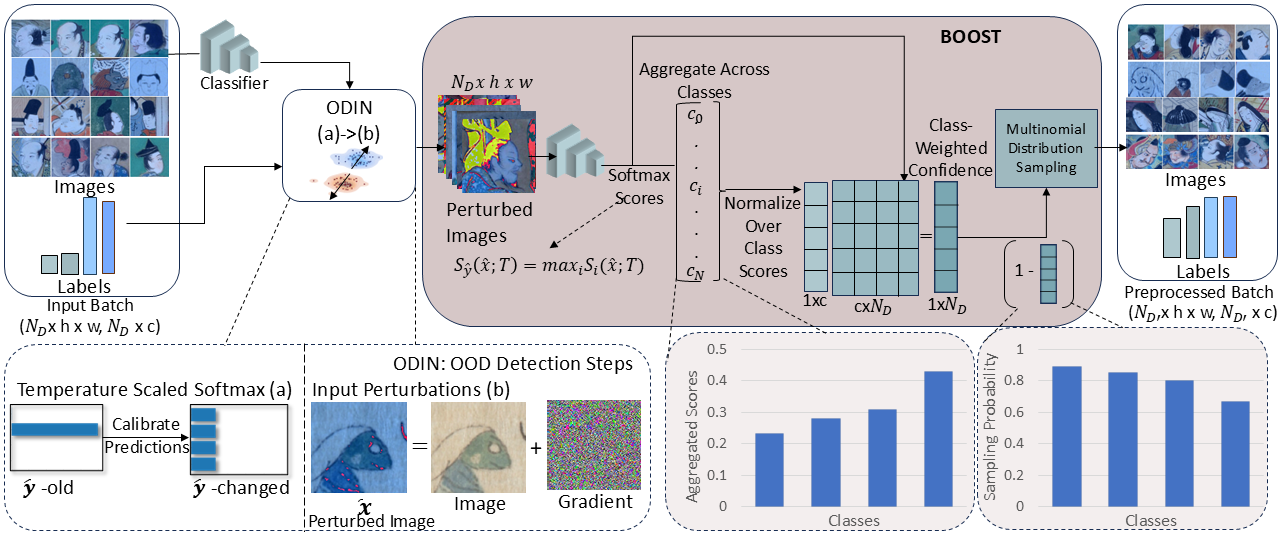}
\caption{The BOOST sampler pipeline sub-samples the training dataset distribution by selecting and resampling input batches to address class imbalance. The ODIN detector first calibrates predictions from passing input batches into a local copy of an image classifier using temperature-scaled softmax and then perturbs inputs to expose challenging cases. Next, the perturbed images are reprocessed to generate softmax scores, which are aggregated and normalized to produce class-weighted confidence scores across the whole dataset. A multinomial sampling process then samples from the flipped the confidence scores to focus on bias mitigation in the training process.}
\label{fig:meth_fig1}
\end{figure*}


\subsection{Asymmetric Training and Testing Setups}
\label{sec:training_testing_4.3}

\textbf{Training Stage:} We employ a backbone classification model (e.g. STSACLF \cite{vijendran23tackling}, which leverages spatial attention blocks in the final classification layers to focus on different levels of context in the images). The proposed BOOST sampler prioritizes ambiguous data early in training, gradually shifting to uniform sampling of rare and common classes to improve imbalanced data classification. After training, our sampler spreads out data over a wider area, resulting in less overlap between samples while showing less overlap between boundary samples of different classes, separating classes more effectively. By training on variations of the training dataset, the model learns from different inter- and intra-class distributions without memorizing the underlying distribution.   

\textbf{Testing Stage:} During testing, the system tracks the converged model and updates both the sampler’s temperature and the model used by the sampler at the end of each epoch. Test dataset images are classified as either ID or OOD, depending on whether the true labels match or differ from the predicted labels. The key difference is that training fine-tunes the model to differentiate within- and between-class samples over multiple epochs, while testing reclassifies samples as OOD or ID in an epoch without altering model parameters. Using the sampler to classify OOD and ID samples during inference with a single retraining epoch keeps in-distribution data and model parameters unaffected. The two class separation enhances class boundaries, reduces redundancies and supports fairness evaluation by measuring inter- and intra-class disparities. Additionally, the temperature scaling parameter reduces any overlap of the two classes from model overfitting on either class due to oversampling images.



\section{BOOST: Bias-Oriented OOD Sampling and Tuning} 
\label{sec:sampler_implement}

To implement the BOOST sampler framework, we focus on targeting misclassified samples to refine the decision boundaries between in-distribution and out-of-distribution data. By targeting these misclassified samples, the sampler compels the model to re-examine and re-weight its understanding of rare or underrepresented classes, thereby enhancing the representational gap between classes. This process encourages the model to adjust its internal representations to better distinguish between ID and OOD samples, improving overall class discrimination.

\subsection{Class-Weighted Adaptive Sampling with Temperature-Scaled Softmax}
\label{boost_sampler5.1}
Our sampler (Figure \ref{fig:meth_fig1}), designed as a versatile black-box module, enables model bias adaptive sampling by identifying challenging or ambiguous samples based on their distance from decision boundaries. It leverages class-aware selection criteria and confidence scores from the ODIN detector’s temperature-scaled softmax to address both inter-class and intra-class imbalances. Without smart and adaptive selection of training data, the model capacity is utilized for frequent samples with small gradients and low confidences, resulting in poor adaptability to less frequent samples.  By treating other classes as out-of-distribution, we maximize the softmax distribution between in- and out-of-distribution classes using temperature scaling. We adapt the ODIN calibration, enabling the sampler to introduce training subsets of varying difficulty and diversity without training objective conflicts. The Temperature-Scaled (TS) softmax score \(S_c(x;T)\) is utilized to calibrate prediction confidences and enhance the variation between scores across multiple classes: \begin{equation}
    S_c(x;T) = \frac{exp(f_c(x)/T)}{\sum^{N_c}_{j=1}exp(f_j(x)/T)},
\end{equation} where \( x \) is the input painting image, \( T \) is the temperature scaling parameter modulating confidence scores, \( N_c \) is the number of classes, and \( c \) is the in-distribution (ID) class for which the score is calculated. Here, \( f_i \) represents the classifier model for class \( i \).  When $T > 1$, the resulting probability distribution is softer and allowing the model to consider underrepresented artistic styles, periods or subjects. It helps distinguish artistic datasets which involve ambiguous or stylistically similar classes.

Since the TS softmax function scales the logits uniformly, the result could be biased towards high-frequency classes in class imbalanced datasets. The temperature scale helps match the maximum of the overall average confidence with the overall accuracy of the model, but requires additional scaling to make the score class-aware. We start the model training with the temperature as 1 to reflect the true training data distribution and then scale the parameter multiplicatively by a constant every 5 epochs. By default, we choose the temperature scale as 5. We choose bigger values when we want the training phase to focus less on rare samples and instead learn a uniform data distribution, which seems to work better for training on smaller data distributions such as PACS.  To account for both class imbalanced and balanced datasets, we use a simple heuristic to scale the scores by the class-wise average to group predictions from all confidence ranges of the same class together. 

At the start of training, the sampler focuses on diverse samples, making overfitting less likely and encouraging the model towards robust, generalizable class representations. As training progresses, the sampler gradually shifts its focus toward more representative and diverse samples, ensuring a comprehensive learning of the data distribution. The temperature gradually increases, softening probability scores, modulating model confidence calibration and selectively guiding exploration toward comprehensive coverage across classes. The flexibility of our sampler allows for its use across different datasets and prediction tasks, making it a valuable tool in model bias adaptive sampling.

\subsection{Bias Mitigation through Inverted Class-Weighted Confidence Scores}
\label{new_metrics5.2}
We first compute the individual scores by keeping the model parameters constant and temporarily changing the model weights per batch to compute the gradients for the gradient sign step of the ODIN model. It adds perturbations to the input so that the maximum of the softmax score is closer to the ID class. We represent the perturbation computation as \begin{equation}
    \mathbf{x}_{\text{adv}} = \mathbf{x} - \epsilon \cdot \text{sign}(\nabla_{\mathbf{x}} S'_{\hat{y}}(\mathbf{x}; \mathbf{T})),
\end{equation} where \( \mathbf{x} \) is the original input painting, \( \epsilon \) is a small positive value representing the perturbation magnitude, \( \text{sign}(\cdot) \) is the sign function, and \(\mathbf{x}_{\text{adv}}\) is the adversarial example (perturbed input). We select $\epsilon$ as 0.05 according the ODIN paper.  The term \( \nabla_{\mathbf{x}} S'_{\hat{y}}(\mathbf{x}; T) \) denotes the gradient of the softmax score \( \mathcal{S'} \) for the target class \( \hat{y} \), calculated from a single forward pass of data to obtain aggregate scores \( S_c \). The gradient of the softmax score is inversely scaled by the precomputed standard deviation of the original training dataset of pretrained models to enable a comparable model response to the per-class OOD and ID samples. If not precomputed, we use the ImageNet standard normal as default. These gradient-based perturbations, from later convolutional model features, disturb the image content features towards being more class representative. This allows for different artistic expressions that come from low-level, stylistic features to represent the same content. The \( T \)  parameter is in the range of [1,1000] similar to the ODIN paper. When the temperature is larger than 1000, we observe that it degrades the classifier's performance. The temperature is selected to modulate the softmax scores to prioritize hard samples earlier in training with sharper probability distributions towards rare classes and a softer distribution towards the end of training to a uniform data representation. We use a temperature scheduler to increase the parameter incrementally during training to change the sampled distribution from the BOOST sampler in predefined epoch intervals. We first explored temperature schedulers that inversely start from 1000 and linearly degrade to 1 with constant steps. We chose not to interpolate the steps across the interval, instead choosing to flatten the temperature after it reaches either bounds to prevent different rates of modulation depending on the training routine and avoiding non-linear effects. We observed that the temperature schedule works best on a piecewise linear schedule, whereas the improvement can degrade or improve at a slower rate for non linear temperature schedules. That is, the piecewise temperature schedule also provides multiple training regimes for controlled difficulty progression where the model is allowed to train until it starts to saturate.

Individual scores result from pre-processing the input with the BOOST-trained model before getting the maximum softmax score. These scores are scaled with the aggregated class-wise scores to form the initial sampling probabilities and are batch normalized to preserve local statistics and relative weighting of samples for that batch.. To incentivize the model to prioritize hard samples, we perform multinomial batch stratification by subtracting the probabilities from 1 and sampling with replacement. By making this adjustment, our method enables underrepresented samples to contribute more to the learning process at the start, reducing learned bias. Our method also introduces a gradient sign step to add perturbations to the input, pushing the softmax score closer to the ID class and refining the model's ability to differentiate between in-distribution and OOD samples. The sampling method employs a multinomial batch stratification with replacement, which incentivizes the model to prioritize rarer classes by adjusting the sampling probabilities accordingly. \begin{equation}
    P(y = c \mid \mathbf{x}) = 1 -\left(\frac{\exp(z_c) \cdot S_c}{\sum_{j} \exp(z_j) \cdot S_j}\right),
\end{equation} where \( z_c \) is the logit for class \( c \), computed as \( z_c = f_{img}(\mathbf{x}_{\text{adv}}) \), with \( f_{img} \) representing the image classification model. The term \( S_k \) denotes the aggregate score for each class \( k \), to account for inter-class relationships in the score calculation. The denominator sums over all classes \( j \), ensuring that the score is class-normalized. 

The BOOST temperature scheduling strategy prevents overconfidence in predictions during later training stages by employing higher temperatures. It encourages a broader sampling range and improves generalization and prevents the model overfitting to a few difficult examples, thereby avoiding the under sampling of rare class samples.

\subsection{Same-Dataset OOD Detection Score (SODC)}
\label{sec:sodc}
To evaluate the class-wise separation and per-class bias reduction, we define the novel Same-Dataset OOD Detection Score (SODC) as the ratio of the number of OOD samples to the total number of samples, where we re-contextualize the ID samples as those correctly classified and OOD samples as the misclassified samples for the incorrect classes. We first use ODIN to preprocess the samples with the input and its corresponding labels as inputs. The softmax scores from the preprocessed inputs form the OOD confidence scores, after which the predictions are matched against the labels to form the count for the OOD and ID samples. The SODC is given by: \begin{equation}
    \text{SODC}_{c} = \frac{\sum_{i=1}^{n} \left( 1(y_i = c) \cdot 1(\hat{y}_i = c) \cdot S_{i,c} \right)}{\sum_{i=1}^{n} \left( 1(y_i = c) + 1(y_i \neq c) \right)},
\end{equation} where \( y_i \) is the true label for the \( i \)-th sample, \( \hat{y}_i \) is the predicted label for the \( i \)-th sample, and \( 1(\cdot) \) is an indicator function that returns 1 if the condition inside is true, and 0 otherwise. \( S_{i,c} \) represents the softmax score of the \( i \)-th logit for class \( c \), and \( n \) is the total number of samples. The total SODC, designed to be sensitive to poor individual class performance, is given by:
\begin{equation}
\text{SODC} = \prod_c {\text{SODC}_c}
\end{equation}
We specify the OOD samples as instances that the classifier incorrectly identifies as belonging to one of the known classes, making them false positives. A high OOD score indicates a stronger ability to filter out these false positives, while a lower score signals poor separation between OOD samples and ID (correctly classified) samples in a class. Our BOOST's pre-processed inputs highly separate the samples from different classes due to their temperature scaling and perturbations. Since the ODIN processed input pushes the input closer to its class's representation space, it has a better gap between the OOD and ID samples making it a good choice for a pseudo ground truth value. The samples that stay misclassified after using BOOST are those proportion of OOD samples for a given class that are highly biased to another class, drastically decreasing the overall OOD score and highlighting group disparities. In case of the control models, i.e. those not trained with BOOST, a random sampler is selected instead of our sampler to get the pseudo ground truth values which lack the advantage of a class-separated representation space.

\subsection{Adapting Resampling Strategies Based on Model Training Dynamics}
In this work, we propose to assess the model bias during training by analyzing low-confidence samples, which often correspond to rare or underrepresented classes. By resampling based on confidence scores, we adjust the training subset distribution to give higher weight to samples from classes with lower original confidence scores. The standard deviation of the sampling distribution is modulated according to the variability in class confidence: it is smaller for classes with consistently high (typical of representative classes) or low (typical of rare classes) confidence, and larger for classes with mixed confidence. This approach allows underrepresented samples to have higher weights while maintaining the original level of variability within each class. The distribution is further refined by scaling individual scores and aligning them with class centroids, which effectively reduces intraclass variation while widening the gap between classes. Additionally, the probabilistic inclusion of samples rather than hard undersampling/oversampling reduces the risk of bias amplification or representation collapse.
	
The debiasing sampler also keeps track of the training data history and stores score information for the entire dataset across epochs. The history enables the sampler to rotate focus across different hard samples, instead of reusing the same ones. This approach also improves model generalization by exposing the model to diverse subsets of rare class examples over time. The sampler implicitly captures an evolving summary of class distributions through the aggregated class-wise softmax profiles, allowing it to approximate the different training data subsets and vary the sampled instances in a way that supports better generalization across underrepresented classes. Monitoring this history allows the sampler to identify areas of the dataset that are underrepresented or overrepresented as training progresses, enabling it to adaptively focus on samples or regions that require more attention. This dynamic and informed resampling process ensures that the sampler can adjust its strategy based on the evolving needs of the model during training. By the end of training, this adaptive sampling approach facilitates a model that achieves more balanced performance across all classes. It effectively balances the trade-off between debiasing the dataset and maintaining overall model performance, enhancing the model's ability to generalize to underrepresented classes without sacrificing accuracy on the full data distribution.

\section{Experiments}
\label{sec:experiments}

\subsection{Datasets}

In the experiment section of our study, we utilized two datasets, KaoKore and PACS, to investigate the influence of style and content biases in image classification tasks.

\textbf{KaoKore:} The KaoKore \cite{tian2020kaokore} dataset consists of cropped images of facial expressions derived from Japanese art. It includes 5552 RGB images, each with a resolution of $256\times 256$ pixels, and is categorized into four distinct classes. However, the dataset exhibits class imbalance, with a notable skew towards images of noble and warrior faces. This imbalance makes KaoKore particularly suitable for experiments focused on mitigating representational bias. The images are highly stylized, offering simple yet distinct representations of facial shapes and hairstyles that show strong correlation across different classes.

\textbf{PACS:} The PACS \cite{li2017deeper} dataset contains images representing objects, humans, and animals depicted across various styles, making it an ideal choice for experiments targeting debiasing either stylistic or content-related biases. This dataset includes 9991 images, distributed across seven categories and four different domains. Our experiments focus on content debiasing where we examined the differences between realistic and non-photorealistic domains, varying in levels of detail, to investigate how biases emerge based on these variations.

\subsection{Bias Detection Metrics}
Accurately measuring and understanding biases in painting classification is critical for ensuring fair and robust model performance across diverse artwork datasets. In this section, we introduce three key metrics—Mean Absolute Bias (MAB), Standard Deviation of Bias (SDB), and Same-Dataset OOD Detection Score (SODC)—each designed to provide a comprehensive assessment of a model’s performance variability and its susceptibility to biases towards subsets of classes.

\textbf{Mean Absolute Bias (MAB):} Mean Absolute Bias (MAB) is a metric that quantifies the average absolute difference between the classwise performance metric of a model and the mean performance metric across all classes. This metric helps to identify the extent of bias a model exhibits towards certain classes. A lower MAB indicates that the model performs more consistently across all classes, while a higher MAB suggests significant deviations in performance across different classes. The MAB of a model is computed by: \begin{equation}
\text{MAB} = \frac{1}{N_c} \sum_{i=1}^{N_c} |\text{PM}_i - \text{Mean PM}|,
\end{equation} where \( \text{PM}_i \) represents the Performance Metric score for class \( i \), and \( \text{Mean PM} \) is the average Performance Metric score across all classes.

\textbf{Standard Deviation of Bias (SDB):} Standard Deviation of Bias (SDB) measures the variability or spread of the bias across different classes. It provides insight into how much the individual classwise biases deviate from the mean bias. A lower SDB indicates that the model's performance is uniformly distributed across classes, while a higher SDB suggests that the model's performance varies significantly from one class to another. The SDB is computed by:
\begin{equation}
\text{SDB} = \sqrt{\frac{1}{N_c} \sum_{i=1}^{N_c} (\text{PM}_i - \text{Mean PM})^2}.
\end{equation}

\begin{table}[]
\caption{Classification performance with different sampling strategies where the sampler selects the training data before every epoch of model training.}
\label{tab:tab2}
\footnotesize
\centering
\begin{tabular}{lllll}
\toprule
Sampling strategy           & Accuracy       & F1             & Recall         & Precision      \\
\midrule
Random sampling             & 77.33          & 76.13          & 76.31          & 81.52          \\
Dynamic random sampling     & 81.49          & 77.70          & 79.18          & 76.65          \\
Stratified sampling         & 78.17          & 72.93          & 75.46          & 73.46          \\
Dynamic stratified sampling & 75.96          & 74.73          & 79.75          & 73.38          \\
ADASYN \cite{kaur2022analyzing} & 63.44          & 27.48          & 32.60          & 27.66          \\
SMOTE \cite{kaur2022analyzing} & 68.82         & 20.97          & 23.53          & 18.98          \\
\textbf{Ours (BOOST sampling)} & \textbf{84.44} & \textbf{79.79} & \textbf{80.49} & \textbf{79.96} \\
\bottomrule
\end{tabular}
\end{table}

\subsection{Quantitative Results}

Our method is fundamentally novel, focusing on bias mitigation in painting classification through an OOD-informed adaptive sampling strategy (BOOST), which is not addressed by traditional or classical OOD methods. But, we compare our method against classical sampling approaches—such as random sampling and stratified sampling— while leaving behind undersampling and oversampling methods since they lead to overfitting the rare samples or underfitting the representative samples without explicitly tackling dataset bias or class imbalance. In Table \ref{tab:tab2}, BOOST shows the best performance across all metrics: highest Accuracy (84.44\%), F1 (79.79\%), Recall (80.49\%), and Precision (79.96\%). Random sampling has the lowest Accuracy (77.33\%) but relatively high Precision (81.52\%). Stratified sampling shows balanced performance but does not excel in any particular metric compared to others. The model frequently misclassifies other classes as noble, but is conservative in its prediction of the class. The model struggles to effectively classify commoners, indicating possible ambiguity between factors common to this class as compared to the other classes. \MV{SMOTE and ADASYN fail at classifying the majority class samples properly which drags down the overall score.}

\begin{table}[t]
\caption{Bias Metrics for BOOST and Control Models on the KaoKore Dataset. The table reports the Performance, Mean Absolute Bias (MAB), and Standard Deviation of Bias (SDB) for Accuracy, F1 Score, Precision, Recall, and Proportion of OOD Samples Detected. The calculated SODC (OOD Score) as detected by the Control models are fine-tuned for 1 epoch on the BOOST sampler to get the expected misclassified samples from the sampler.}
\label{tab:bias_metrics}
\centering
\footnotesize
\begin{tabular}{llcccccc}
\toprule
\multirow{2}{*}{\textbf{Dataset}} & \multirow{2}{*}{\textbf{Metric}} & \multicolumn{3}{c}{\textbf{Ours (BOOST)}} & \multicolumn{3}{c}{\textbf{Control}} \\

                                  &                                  & \textbf{Score (\%)} & \textbf{MAB (\%) $\downarrow$} & \textbf{SDB (\%) $\downarrow$}   & \textbf{Score (\%)} & \textbf{MAB (\%) $\downarrow$}   & \textbf{SDB (\%) $\downarrow$}   \\
\midrule
\multirow{5}{*}{KaoKore} 
    & \textbf{Accuracy $\uparrow$}  & \textbf{84.44} & \textbf{2.94}   & \textbf{3.51}   & 79.51 & 3.605 & 4.48\\
    & \textbf{F1 Score $\uparrow$}  & \textbf{79.79} & \textbf{9.24}   & \textbf{11.09}  & 73.43 & 11.23  & 13.53  \\
    & \textbf{Precision $\uparrow$} & 80.49 & \textbf{6.20}   & \textbf{8.29}   & \textbf{82.48} & 8.71   & 9.41   \\
    & \textbf{Recall $\uparrow$}    & \textbf{79.96} & \textbf{10.84}  & \textbf{12.96}  & 69.34 & 14.60  & 17.06  \\
    & \textbf{SODC $\uparrow$} & \textbf{1.84} & \textbf{2.73} & \textbf{2.84} & 0.44 & 7.64 & 9.94 \\
\midrule
\multirow{5}{*}{PACS Content} 
    & \textbf{Accuracy $\uparrow$}  & \textbf{94.8} & \textbf{0.29}   & \textbf{0.34}   & 91.56 & 0.50   & 0.53  \\
    & \textbf{F1 Score $\uparrow$}  & \textbf{94.21} & \textbf{0.82}   & \textbf{0.97}  & 90.88 & 1.98  & 2.31  \\
    & \textbf{Precision $\uparrow$} & \textbf{94.88} & \textbf{0.95}   & \textbf{1.18}   & 91.06 & 2.97   & 3.29   \\
    & \textbf{Recall $\uparrow$}    & \textbf{94.05} & \textbf{1.87}  & \textbf{1.92}  & 91.24 & 3.00  & 3.72  \\
    & \textbf{SODC $\uparrow$} & \textbf{0.472} & \textbf{0.89} & \textbf{0.01} & 0.43 & 1.46 & 1.63 \\
\bottomrule
\end{tabular}
\end{table}

Table \ref{tab:bias_metrics} reports the Mean Absolute Bias (MAB) and the Standard Deviation of Bias (SDB) for each performance and OOD detection metric, providing insight into the models' performance consistency and susceptibility to bias across different classes within the KaoKore dataset. It shows the results of detecting the proportion of OOD samples during the training process of the STSACLF model for 20 epochs. The BOOST sampler helps the model reach a better performance against the random sampling strategy for all three metrics under accuracy, F1, recall and SODC. Our model reached a worse precision performance as a result of the tradeoff in performance from encouraging the model to predict difficult samples that it’s unconfident with.
The BOOST trained model exhibits a lower Mean Absolute Bias and Standard Deviation of Bias in accuracy compared to the Control model. This suggests that our model is more consistent in its performance across different classes, making it less prone to favoring or penalizing specific classes. This consistency is crucial for ensuring that no particular class is underrepresented or overrepresented, thereby promoting fairness in classification. When detecting OOD samples, BOOST vastly outperforms the control, showing similar misclassifications across the classes as compared to the Control where the samples are commonly misclassified into the Noble class due to high overlap in the representation space as shown in the qualitative experiments findings.  Additionally, the table showcases our BOOST sampler's strengths in class-wise performance by comparing the metrics against the control where the model is trained without any sampling strategies. It has lower deviations across the precision entries as compared to recall to hint that the model prefers less confusion in its prediction with fewer false positives. Inconsistent recall from higher bias suggests some classes are harder to capture, likely due to subtle variations or insufficient training examples.

\begin{table}[t]
\centering
\caption{Comparison of BOOST against the control performance across varying long-tailed distributions on the PeopleArt dataset.}
\resizebox{\textwidth}{!}{%
\begin{tabular}{lllllllll}
\hline
\multirow{2}{*}{PeopleArt dataset long tail distribution extent} & \multicolumn{4}{l}{BOOST Sampler}                                         & \multicolumn{4}{l}{Random Sampler (Control)} \\
                                                                  & Accuracy         & F1               & Recall           & Precision        & Accuracy  & F1       & Recall   & Precision  \\ \hline
original                                                          & \textbf{48.24\%} & \textbf{43.01\%} & \textbf{42.67\%} & \textbf{46.21\%} & 46.84\%   & 39.10\%  & 39.79\%  & 41.18\%    \\
Pareto (scale = -0.5)                                             & \textbf{45.18\%} & \textbf{41.84\%} & \textbf{40.72\%} & \textbf{43.01\%} & 43.99\%   & 37.77\%  & 38.56\%  & 40.79\%    \\
Pareto (scale = -0.2)                                             & \textbf{55.7\%}  & \textbf{52\%}    & \textbf{50.21\%} & \textbf{53.9\%}  & 52.67\%   & 35.16\%  & 37.13\%  & 38.87\%    \\
Pareto (scale = 0)                                                & \textbf{42.62\%} & \textbf{39.98\%} & \textbf{40.83\%} & \textbf{39.15\%} & 39.02\%   & 26.43\%  & 29.47\%  & 26.99\%    \\ \hline
\end{tabular}
}
\label{tab:long_distro}
\end{table}

\MV{To evaluate the performance of our sampler against long tail training distributions, we use the PeopleArt dataset and extract the ranked number of samples per class. We then plot a pareto curve to approximate a long tailed distribution's exponential function trait with different scales and positions for the tail start. Then we select evenly spaced points scaled up by the max rank class on the pareto curve with varying scales and the default location from the scipy library. The class samples are uniformly sampled to meet the number of samples at each interpolated rank. If the class has less samples than the parent curve point, it is randomly oversampled to make the number. We train our model against the samples (in the resampled dataset) selected from the random sampler and BOOST sampler respectively. We notice in Table \ref{tab:long_distro} that the best performance is achieved by our BOOST sampler and on the original dataset that doesn't lack information. The BOOST sampler reaches a more classwise unifom performance, getting similar ranges of F1, recall, precision to the accuracy. Both the samplers tend to get higher precision over recall for the long tailed datasets, indicating that the model makes more conservative predictions, but doesn't learn for the whole range of diverse outputs in classes.}

\subsection{Qualitative Results}

We visualize the samples using UMAP's underlying fuzzy simplicial set to get the neighborhood connections from its weighted K-NN graph since the distances between the nodes don't have meaningful interpretations. We use this neighborhood connection graph to encode local and global connectivity for visualization intra class distances through contours and samples with high inter class connectivity through thresholding the sum of the edge weights to other classes' samples.

\begin{figure}[!ht]
 
        \includegraphics[width=\textwidth]{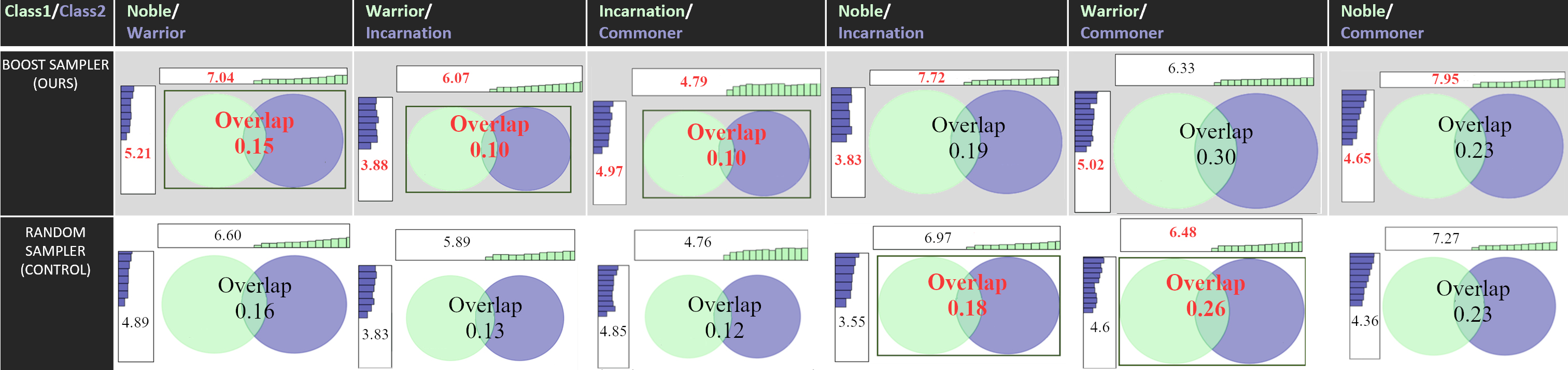} 
        \caption[short for lof]{In the figure, we use the following color scheme: The green boxes on the venn diagrams show which pairwise overlap is smaller for the two cases, where the} light green and dark blue colors represent distinct classes. The green numbers indicate the larger average intra-class distances and vice-verse for the red numbers. The red text indicates the better results. We visualize inter-class overlap and intra-class density distributions for various class pairs using 2D painting dataset projections. The 2D projections show inter-class overlap and intra-class density for class pairs, comparing embeddings from our sampler-trained model with the control. Venn diagrams illustrate inter-class separability, with overlap indicating normalized intersection areas. Insets provide 1D histograms of intra-class distances.

\label{fig:task_complexity}
\end{figure}

Figure  \ref{fig:task_complexity} shows 2D UMAP (Uniform Manifold Approximation and Projection) \cite{mcinnes2018umap}  embeddings visualizing task complexity within and between classes in the KaoKore dataset after training. We qualitatively assess inter-class and intra-class overlap for our sampler versus a random sampler using the STSACLF classifier. We illustrate pairwise class overlap through Venn diagrams and intra-class spread via 1D distance histograms. To this end, we first the feature embeddings for the model trained from different sampler strategies and compute the pairwise cosine distances. The cosine distances indicate the feature similarities across and within classes. We then get the average cosine distances within a class by filtering those against itself. Similarly, we get the mean for the feature distances against other classes and threshold these averages to get different overlap levels. If the overlap is low, the classes are well separated and vice-versa. We use these distances to get the frequency of features per class with different levels of separability. The class boundaries are outlined, and the overlap is quantified as the proportion of points of one class within the boundary of another. Venn diagrams display the extent of inter-class overlap, while histograms reveal intra-class compactness. \MV{The numbers on the venn diagram are calculated as the proportion of those samples between pairwise classes with high connectivity against the lower connectivity samples. The numbers on the histogram refer to the average histogram bin height for the same class pairwise cosine distance of the embeddings. This bin height is proportional to the number of samples that overlap across different regions of the class.} Our sampler achieves better intra-class clustering and lower inter-class overlap (average overlap: 0.18 for our sampler vs. 0.2 for control).

\begin{figure}[ht]

\centering
    \begin{subfigure}{0.45\columnwidth}
        \centering
        \includegraphics[width=.85\textwidth]{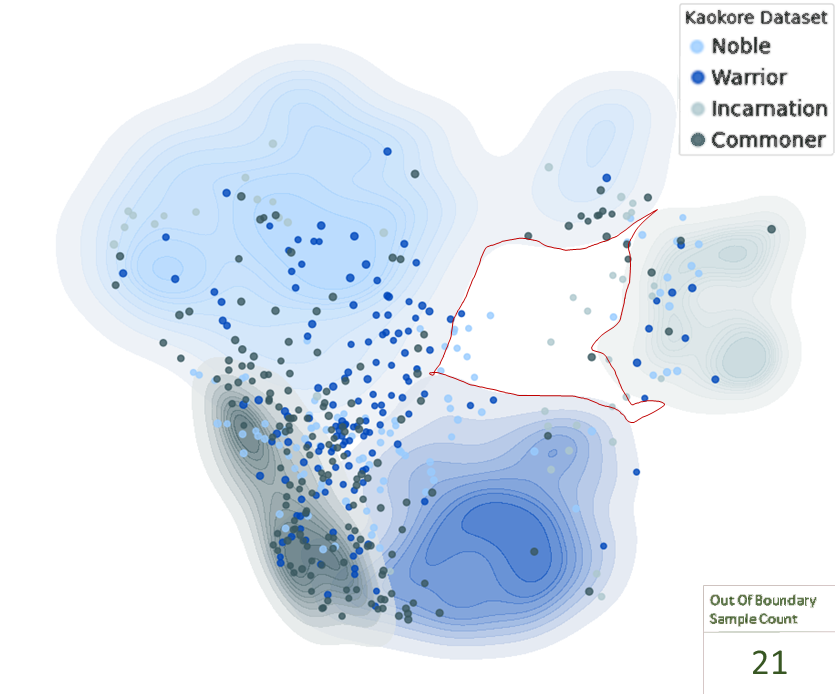} 
        \caption{BOOST Sampler}
        \label{fig:image1}
    \end{subfigure}
    \hfill
    \begin{subfigure}{0.45\columnwidth}
        \centering
        \includegraphics[width=.85\textwidth]{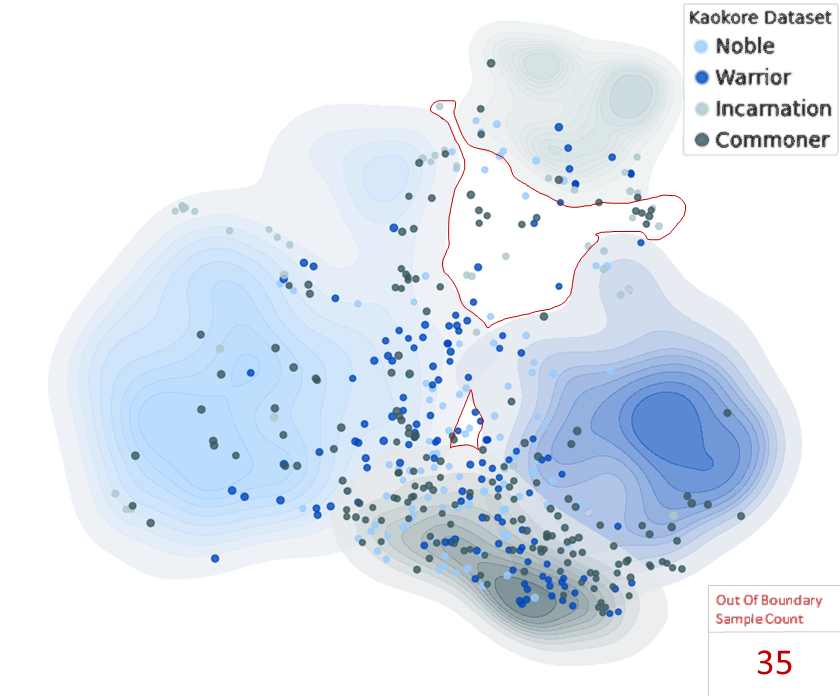} 
        \caption{Random Sampler}
        \label{fig:image2}
    \end{subfigure}
 \caption[short for lof]{Visualizations for embeddings with UMAP projections on the underlying K-Nearest-Neighbors graph to measure data connectivity. The dot markers are class-wise data that have 80\% or more of the nearest neighbors connected to samples of other classes. The underlying contours represent the connection density of embedding distributions from common to rare samples based on neighbor counts. The darker regions represent samples with high number of connections and vice versa.  The space enclosed by the red boundaries contain the out of boundary samples with the total count in the bottom right box. This box highlights the better sampler, BOOST, in green with the opposite case in red.}]
\label{fig:samp_distro}

\end{figure}

Figure \ref{fig:samp_distro} compares BOOST and Random Sampler strategies using UMAP projections and its K-Nearest Neighbors (K-NN) graph to assess data connectivity. The K-NN graph is used in UMAP for approximating the low dimension representation of the data distribution, preserving the pointwise connectivity in local and global contexts. The visualization highlights the degree of data separation from dense region connections within the same class, as well as poorly separated areas consisting of ambiguous samples connecting different classes. We select UMAP because it balances local and global clusters while preserving variance from a PCA-initialized graph. The PCA data point position initialization is widely used to provide an efficient starting point for optimizing low-dimensional topological representations. The UMAP on its own provides a visualization approximating the local and global structure of the dataset, but is insufficient for our purpose of visualizing the inter and intra class distances since any edge distances are not meaningful as absolute measurements. For each K-NN node, we calculate the proportion of connections to the same class versus other classes, using a threshold (0.8) to mark high cross-class connectivity with 'o' markers. Points exceeding this threshold are marked to indicate inter-class links. A density plot of 2D projections further visualizes intra-class connectivity, with contours clustering in dense areas and spreading in sparse regions. The anomalous samples are those found outside the contours and are enclosed in the red boundary line. These samples are found at outside the class wise contours, showing their dissimilarity with the class-wise samples. BOOST demonstrates tighter intra-class connectivity and better separation, with its anomalous samples closer to the class wise contours..
The Random Sampler shows widespread inter-class mixing with weaker distinctions due to uniform contouring, while BOOST creates distinct high-connectivity regions, clearly defining class boundaries. Fewer marked points (21) in BOOST indicate better data separation, with reduced inter-class connections. In contrast, the Random Sampler places neighboring points from other classes in sparse regions (35), highlighting poor separation. \MV{Figure \ref{fig:samp_distro} also provides justification to the overlapping proportions. The BOOST sampler on average gets a lower interclass overlap than the Random sampler, but the Noble/Incarnation and Warrior/Commoner classes have higher overlaps corresponding to more samples from the high interclass samples of other classes in the class contours.  There are more marked points from the Warrior class in the Commoner class contours for the BOOST sampler compared to the Random sampler. There are more marked samples from each others class for the Noble and Incarnation classes in each others regions for the BOOST sampler as compared to the Random sampler. The class pairs with lower interclass overlap have more samples with high connectivity to the opposite class. It serves as a measure of ambiguity between classes. The samples that contribute more to the overlap proportions are those ambiguous samples from the classes with more samples such as Noble or Warrior that are statistically more likely to have ambiguous samples with other classes, especially if there are features common across all the classes. The Kaokore dataset has simple facial features but detailed backgrounds or accessories. These are commonly shared features across classes, leading to this ambiguity.}

\begin{figure}[!ht]
    \centering

    \begin{subfigure}{\columnwidth}
        \centering
        \includegraphics[width=0.5\linewidth]{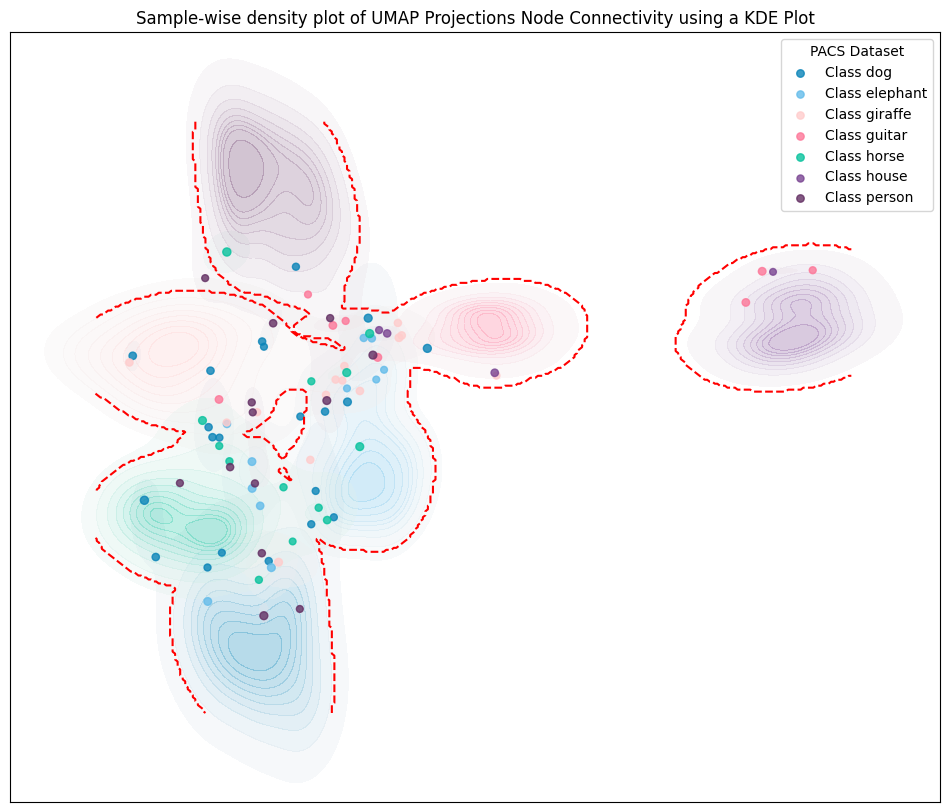}
        \subcaption{PACS Content Embeddings Connectivity Visualization}
        \label{fig:subfig1}
    \end{subfigure}
    \begin{subfigure}{0.48\columnwidth}
        \centering
        \includegraphics[width=0.8\linewidth]{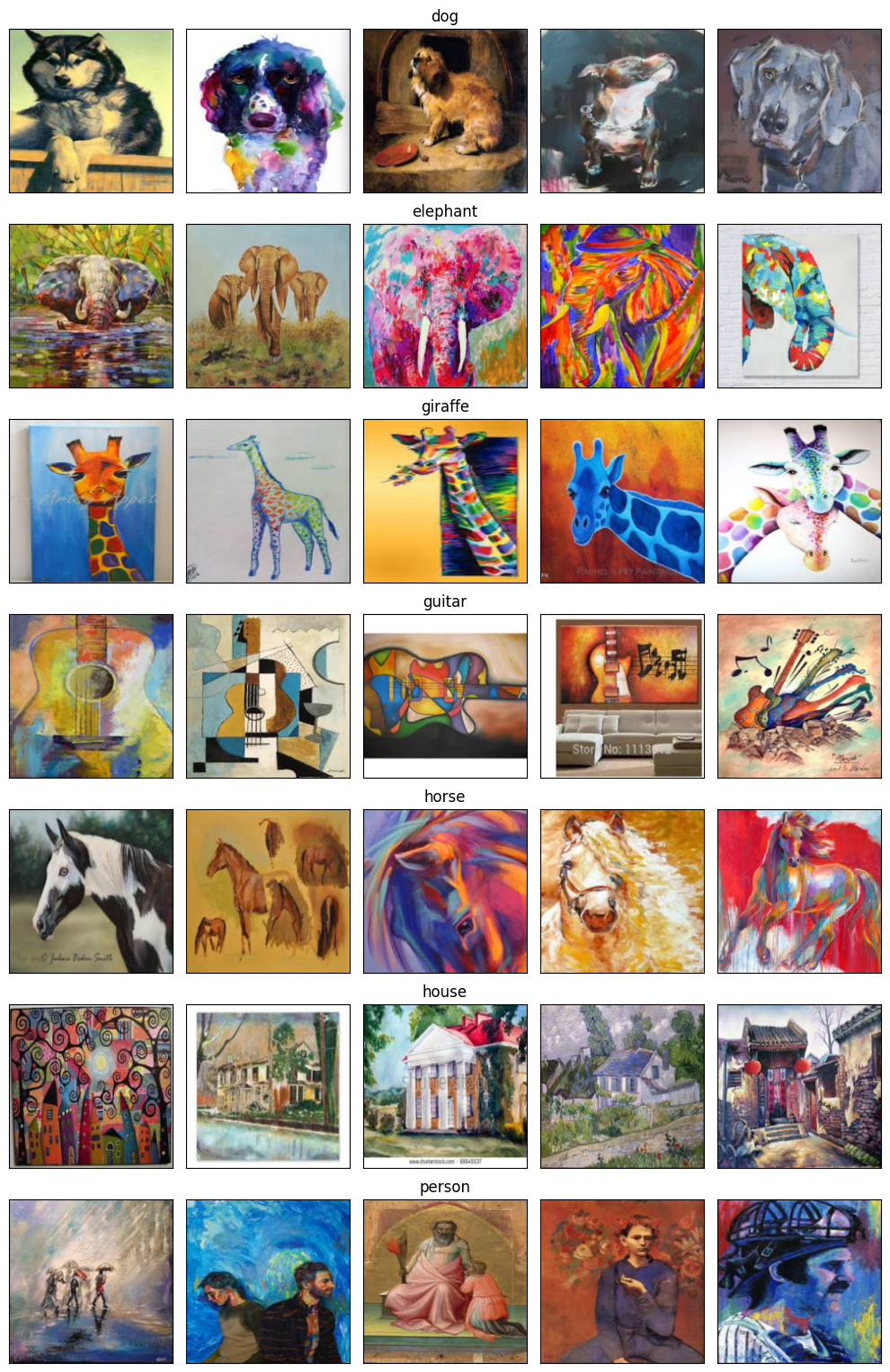}
        \subcaption{Grid-wise Visualization of a Sample Subset of Highly Inter-class Connected Samples. }
        \label{fig:subfig2}
    \end{subfigure}
    \begin{subfigure}{0.45\columnwidth}
        \centering
        \includegraphics[width=0.8\linewidth]{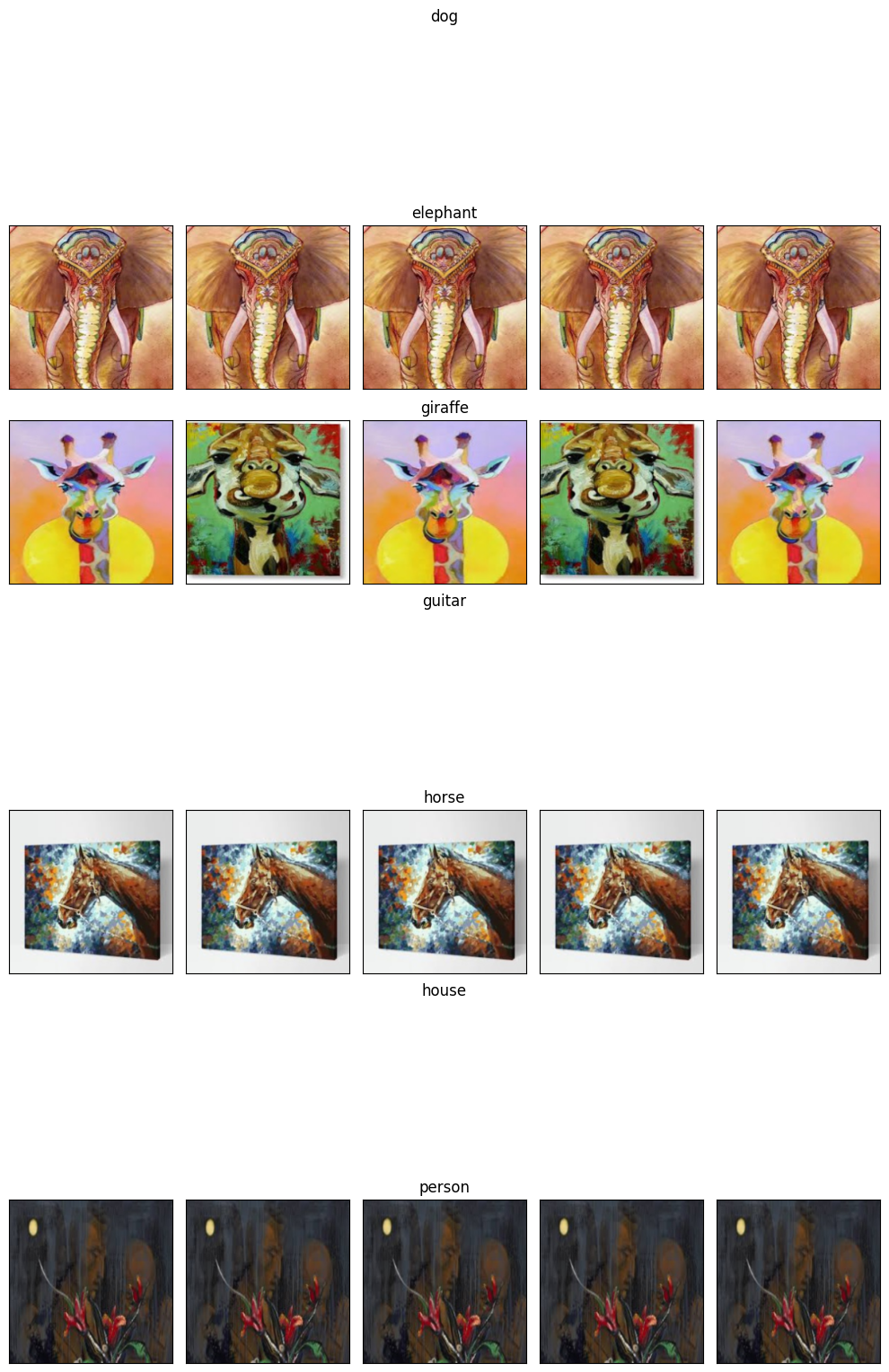}
        \subcaption{Grid-wise Visualization of a Sample Subset of OOD Samples in the Red Boundary Region. }
        \label{fig:subfig3}
    \end{subfigure}
    \caption{Visualization of the samples with high connectivity and anomalous samples in the BOOST sampler trained model embeddings. For the grids, each row is prefixed with the class name. The samples repeat on exhaustion of the listed samples for the class. Otherwise, the samples for the class are picked at random from the two sets for each row of the grid. If there are no samples for a class, we leave the row empty.}
    \label{fig:connectivity_sample_grids}
\end{figure}

We explore a focused sampling for Figure \ref{fig:connectivity_sample_grids} using the PACS dataset to showcase the learned model behavior for different types of content. It provides a comprehensive visualization of the BOOST sampler interaction with the content embeddings in the PACS Dataset from the connectivity structure and the selected sample grid. The classes dog and elephant overlap because of their large ear flaps visualized in the paintings. Similarly, classes such as horses and giraffes are placed close to each other due to their structural similarity, or humans and houses due to their close association. 

Next, the left grid shows high inter-class inter-connectivity, i.e. the sample points with markers over the contours. The samples in the left grid exemplify the types of challenging instances the sampler prioritizes. As noted, stylistic similarities (like vibrant palettes or abstract forms seen in the giraffe, elephant, and some dog paintings) and part-wise resemblances create this ambiguity. For instance, the heavily stylized, colorful examples are found in these high-connectivity areas because their visual features might resonate with multiple class centroids in the embedding space, making classification difficult. These samples show overlap with other classes due to their palette choice (such as the giraffe rainbow coloring and that of dogs, elephants, etc.) or if their parts vaguely resemble another class (such as the three elephants in the second row looking slightly like a dog face). Thus, the model struggles with stylistic similarities that make images of different contents ambiguous or those images whose partwise composition resembles another class. 

Furthermore, the right grid showcases samples where the foreground and background palettes bleed into each other (for the elephant and giraffe) or if the overall dynamic range is limited (for the person). For classes like 'Elephant' and 'Giraffe', which could be considered rare in terms of absolute sample count or stylistic diversity in a dataset compared to everyday objects/subjects, this grid highlights a specific representation: highly stylized and vibrantly colored artistic renderings. The images exhibit significant foreground-background bleed, where the subjects are rendered with palettes that merge with or are indistinguishable from the background colors. This stylistic choice, while visually striking, can make object segmentation and recognition difficult for a model relying on clearer boundaries or more typical color schemes. The sampler targets these examples because they represent a specific visual domain shift within the class, pushing the model to generalize beyond conventional photographic or less abstract artistic styles. Similarly, for the 'Horse' and 'Person' classes, the anomalous images often feature darker palettes, painting strokes that obfuscate discriminative features, and potentially limited dynamic range. For instance, the person images are consistently dark, with subjects often merging into shadowy or richly colored backgrounds. These qualities challenge the model's ability to discern subjects in low-contrast environments or generalize across different lighting conditions and artistic interpretations. Thus, the right grid illustrates how the sampler identifies and prioritizes samples not just from areas of inter-class overlap (as seen in the left grid), but also samples that represent visually atypical, stylized, or challenging instances within individual class distributions. By focusing on these less common visual manifestations (be it vibrant bleeding, speckled textures, or low-light blending), the sampler encourages the model to become more robust to diverse artistic styles and visual complexities that deviate from standard or more frequent data examples, regardless of the overall class frequency in the dataset.

\begin{figure}[!ht]
    \centering

    \begin{subfigure}{0.45\columnwidth}
        \centering
        \includegraphics[width=.7\linewidth]{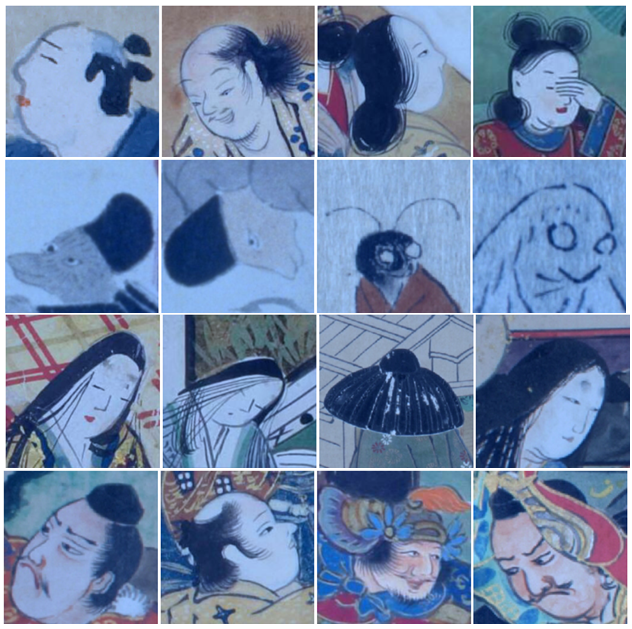}
        \subcaption{Most Diverse Samples.}
        \label{fig:subfig1}
    \end{subfigure}
    \begin{subfigure}{0.45\columnwidth}
        \centering
        \includegraphics[width=.7\linewidth]{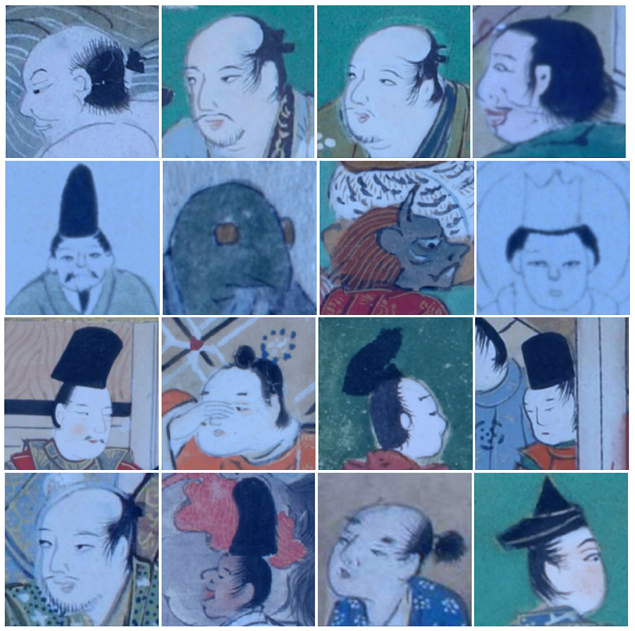}
        \subcaption{Least Diverse Samples.}
        \label{fig:subfig2}
    \end{subfigure}
    \caption{Grid of samples sorted according to our BOOST sampler sampling score along the columns. (a) Visualizes the most diverse samples per class with the rows indicating samples from different classes, while the columns show samples with a decreasing order of sampling scores from left to right. (b) Visualizes the least diverse samples per-class by displaying samples with an increasing order of sample scores from left to right instead.}
    \label{fig:diverse_sampling}
\end{figure}

\MV{Furthermore, the BOOST sampler shows remarkable capability in producing more diverse set of samples compared to traditional random sampling techniques. For each class in the KaoKore dataset, we track the sampling scores and confidence distributions assigned to the test samples by the BOOST sampler. To better understand the effect of our BOOST sampler, we visualize the samples with our highest and lowest sorted sampling scores. As shown in Figure \ref{fig:diverse_sampling}, we organize these samples in a grid, where each row represents samples from different classes, and the columns correspond to the decreasing and increasing order of the BOOST sampling score. The sampling score is inversely correlated to the model confidence scores, with the confidence derived from the analysis and interpretation of the image's semantic features. For example, a KaoKore image with a highly abstract style, obscured facial features due to artistic rendering, an unusual pose, or attributes that blend characteristics of multiple classes will likely present ambiguous semantic features to the model. The model cannot confidently assign it to a single class based on its learned feature representations. The grid displays a range of facial expressions and styles in classes like noble and warrior, showing distinct sampling score separation, while visually variable classes like commoner show less distinction. These are the samples the sampler will give a high score to. The lower confidence scores are more likely to correspond to OOD samples, particularly where artistic abstractions, like stylized features or incomplete facial expressions, make classification challenging. In contrast, the samples with higher scores are more confidently classified as ID, typically featuring clearer and more characteristic facial details that align closely with the learned representations of the KaoKore classes.}

\section{Conclusion}
\label{sec:conclusion}

In this work, we presented a novel approach to mitigating biases in deep learning models for painting classification by leveraging OOD methods. Our OOD-informed adaptive sampling targets biases in class-imbalanced datasets, especially those dominated by certain artistic styles. By dynamically adjusting the temperature scaling and sampling probabilities, our method effectively promotes a more equitable representation of all classes, including those that are underrepresented or ambiguous. The proposed BOOST sampler is versatile and can be integrated as a plug-and-play module with various deep learning architectures, making it a valuable tool for researchers and practitioners aiming to develop unbiased AI systems in the art domain. Our experimental results on the KaoKore dataset demonstrated the efficacy of the proposed method, achieving a classification accuracy of 84.44\% and an F1 score of 79.79\%. Moreover, our sampler significantly reduced bias across various performance metrics, showing the lowest MAB and SDB when compared to control models. These results indicate that our method not only improves overall model performance but also enhances fairness in the classification of paintings by minimizing the impact of dataset biases.

Currently, the BOOST sampler primarily focuses on mitigating biases related to style within the dataset. Future work will extend sampler-adaptive training in two key directions. First, we will adapt samplers for diverse downstream tasks by nudging embeddings toward task-specific objectives. Selecting data subsets using instance-wise information from model logits facilitates image reconstruction, segmentation, or classification across parallel image priors \cite{huijben2020learning}, while leveraging neighbourhood distances and embedding similarities enhances embedding contrast, making it transferable across datasets and tasks \cite{wu2021conditional}. Second, we will explore multi-conditionals or multi-modal samplers to refine fine-grained representations and semantic context alignment. Conditional strategies improve classification performance and consolidate multi-modal representations by aligning visual and conditional features \cite{byun2022grit, huang2024neighbor}. Incorporating loss functions to increase sampling score variance further enhances sample quality and embedding alignment. These extensions will further improve the model's ability to handle bias across different dimensions of artistic representation.

\section*{Acknowledgment}
This research is supported in part by the EPSRC NortHFutures project (ref: EP/X031012/1).

\bibliographystyle{elsarticle-num} 
\bibliography{citations}

\end{document}